# Deep learning enhanced mobile-phone microscopy

Yair Rivenson[1,2,3]†, Hatice Ceylan Koydemir[1,2,3]†, Hongda Wang[1,2,3]†, Zhensong Wei[1], Zhengshuang Ren[1], Harun Günaydın[1], Yibo Zhang[1,2,3], Zoltán Göröcs[1,2,3], Kyle Liang[1], Derek Tseng[1], Aydogan Ozcan[1,2,3,4]*

**Affiliations:**

[1]Electrical and Computer Engineering Department, University of California, Los Angeles, CA, 90095, USA.

[2]Bioengineering Department, University of California, Los Angeles, CA, 90095, USA.

[3]California NanoSystems Institute (CNSI), University of California, Los Angeles, CA, 90095, USA.

[4]Department of Surgery, David Geffen School of Medicine, University of California, Los Angeles, CA, 90095, USA.

*Correspondence: ozcan@ucla.edu

Address: 420 Westwood Plaza, Engineering IV Building, UCLA, Los Angeles, CA 90095, USA

Tel: +1(310)825-0915

Fax: +1(310)206-4685

†Equal contributing authors.

**Abstract**

Mobile-phones have facilitated the creation of field-portable, cost-effective imaging and sensing technologies that approach laboratory-grade instrument performance. However, the optical imaging interfaces of mobile-phones are not designed for microscopy and produce spatial and spectral distortions in imaging microscopic specimens. Here, we report on the use of deep learning to correct such distortions introduced by mobile-phone-based microscopes, facilitating the production of high-resolution, denoised and colour-corrected images, matching the performance of benchtop microscopes with high-end objective lenses, also extending their limited depth-of-field. After training a convolutional neural network, we successfully imaged various samples, including blood smears, histopathology tissue sections, and parasites, where the recorded images were highly compressed to ease storage and transmission for telemedicine applications. This method is applicable to other low-cost, aberrated imaging systems, and could offer alternatives for costly and bulky microscopes, while also providing a framework for standardization of optical images for clinical and biomedical applications.

**Introduction**

Optical imaging is a ubiquitous tool for medical diagnosis of numerous conditions and diseases. However, most of the imaging data, considered the gold standard for diagnostic and screening purposes, are acquired using high-end benchtop microscopes. Such microscopes are often equipped with expensive objectives lenses and sensitive sensors, are typically bulky, must be operated by trained personnel, and require substantial supporting infrastructure. These factors potentially limit the accessibility of advanced imaging technologies, especially in resource-limited settings. Consequently, in recent years researchers have implemented cost-effective, mobile microscopes, which are often based on off-the-shelf consumer electronic devices, such as smartphones and tablets.[1] As a result of these research efforts, mobile-phone-based microscopy has demonstrated promise as an analytical tool for rapid and sensitive detection and automated quantification of various biological analytes as well as for the imaging of, e.g., pathology slides[1–11].

In this research, we attempted to bridge the gap between cost-effective mobile microscopes and gold standard benchtop microscopes in terms of their imaging quality. An important challenge in creating high-quality benchtop microscope equivalent images on mobile devices stems from the motivation to keep mobile microscopes cost-effective, compact and light-weight. Consequently, most mobile microscope designs employ inexpensive, often battery-powered illumination sources, such as light-emitting diodes (LEDs), which introduce colour distortions into the acquired images. Furthermore, mobile microscopes are usually equipped with low numerical apertures (NAs) also containing aberrated and often misaligned optical components, which add further distortions into the acquired images at the micro-scale. Although the lenses of mobile-phone cameras have advanced significantly over the last several years, large volume fabrication techniques are employed in the moulding and assembly of these plastic lenses, which creates random deviations for each mobile camera unit compared with the ideal optical design and alignment. Some of these distortions also vary to some degree as a function of time and usage, due to, e.g., the battery status of the mobile device and the illumination unit, the poor mechanical alignment precision of the sample holder, and the user experience. Furthermore, since most optoelectronic imagers found in consumer electronic devices including smartphones have been optimized for close and mid-range photography rather than microscopy, they also contain built-in design features such as varying micro-lens positions with respect to the pixels, which create additional spatial and spectral distortions for microscopic imaging. Finally, since mobile-phone cameras have small pixel sizes (on the order of 1-2 µm) with a very limited capacity of a few thousand photons per pixel, such mobile imagers also have reduced sensitivity. In contrast, high-end benchtop microscopes that are used in medical diagnostics and clinical applications are built around optimized illumination and optical pick-up systems with calibrated spectral responses, including diffraction-limited and aberration-corrected objective lenses and highly-sensitive CCDs (charged-coupled devices) with large pixels.

Here, we describe the substantial enhancement of the imaging performance of a bright-field mobile-phone based microscope using deep learning. The mobile microscope was implemented using a smartphone with a 3D-printed optomechanical attachment to its camera interface, and the image enhancement and aberration correction were performed computationally using a deep convolutional neural network (see Fig. 1, Supplementary Fig. 1, and the Methods section). Deep learning[12] is a powerful machine learning technique that can perform complex operations using a multi-layered artificial neural network and has shown great success in various tasks for which data are abundant[13–16]. The use of deep learning has also been demonstrated in numerous biomedical applications, such as diagnosis[17,18], image classification[19], among others[20–24]. In our method, a supervised learning approach is first applied by feeding the designed deep network with input (smartphone microscope images) and labels (gold standard benchtop microscope images obtained for the same samples) and optimizing a cost function that guides the network to learn the statistical transformation between the input and label. Following the deep network training phase, the network remains fixed and a smartphone microscope image input into the deep network is rapidly enhanced in terms of spatial resolution, signal-to-noise ratio, and colour response, attempting to match the overall image quality and the field of view (FOV) that would result from using a 20× objective lens on a high-end benchtop microscope. In addition, we demonstrate that the image output by the network will have a larger depth of field (DOF) than the corresponding image acquired using a high-NA objective lens on a benchtop microscope. Each enhanced image of the mobile microscope is inferred by the deep network in a non-iterative, feed-forward manner. For example, the deep network generates an enhanced output image with a FOV of ~0.57 mm$^2$ (the same as that of a 20× objective lens), from a smartphone microscope image within ~0.42 s, using a standard personal computer equipped with a dual graphics-processing unit. This deep learning-enabled enhancement is maintained even for highly compressed raw images of the mobile-phone microscope, which is especially desirable for storage, transmission and sharing of the acquired microscopic images for e.g., telemedicine applications, where the neural network can rapidly operate at the location of the remote professional who is tasked with the microscopic inspection of the specimens.

Employing a deep network to learn the statistical transformations between the mobile and optimized benchtop microscope images enabled us to create a convoluted mapping between the two imaging instruments, which includes not only a spatially and spectrally varying distorted point-spread function and the associated colour aberrations, but also a non-uniform space warping at the image plane, introduced by the inexpensive mobile-phone microscope. Unlike most image enhancement methods, this work does not consider physical degradation models during the image formation process. Such image degradation models are in general hard to estimate theoretically or numerically, which limits the applicability of standard inverse imaging techniques. Moreover, even if such a forward model could be estimated, there are almost always unknown and random deviations from it due to fabrication tolerances and alignment imperfections that are unfortunately unavoidable in large scale manufacturing. Instead of trying to come up with such a forward model for image degradation, the deep neural network learns how to predict the benchtop microscope image that is most statistically likely to correspond to the input smartphone microscope image by learning from experimentally-acquired training images of different samples.

We believe that this presented approach is broadly applicable to other low-cost and aberrated microscopy systems and could facilitate the replacement of high-end benchtop microscopes with mobile and cost-effective alternatives, and therefore might find numerous applications in global health, telemedicine and diagnostics related applications. Furthermore, this deep learning enabled image transformation and enhancement framework will also help with the standardization of optical images across various biomedical imaging platforms, including mobile microscopes that are being used for clinical and research applications, and might reduce potential discrepancies in microscopic investigation and diagnostic analysis of specimens, performed by medical professionals.

**Results**

Schematics of the deep network training process are shown in Fig. 2. Following the acquisition and registration of the smartphone and benchtop microscope images (see Methods section), where the benchtop microscope was equipped with a 20× objective lens (NA=0.75), the images were partitioned into input and corresponding label pairs. Then, a localized registration between input and label was performed using pyramid elastic registration to correct distortions caused by various aberrations and warping in the input smartphone microscope images (see Methods section, Fig. 2 and Supplementary Fig. 2). These distortion-corrected images were divided into training and validation sets. The validation set prevented the network from overfitting to the training set, as we used the model achieving the minimal target cost function for the validation set to fix the network parameters. An independent testing set (which was *not* aberration-corrected) enabled us to blindly test our network on samples that were not used for the network training or validation.

The training dataset was generated by partitioning the registered images into 60×60 pixel and 150×150 pixel patch images (with 40% overlap), from the distorted smartphone and the gold standard benchtop microscope images, respectively (the numbers of training patches and the required training times for the different samples are provided in Table 1). We trained multiple networks, corresponding to multiple types of pathology samples (such as stained lung tissue, Papanicolaou (Pap) smear, and blood smear samples), while maintaining the exact same neural network architecture. Following the training of the deep networks (see Table 1 for details), the networks remained fixed and were used to blindly test samples from different pathology slides.

First, we applied our deep learning framework to Masson's-trichrome-stained lung tissue. A representative result is shown in Fig. 3, which demonstrates the ability of the deep network to restore spatial features that cannot be detected in the raw smartphone microscope image due to various factors including spatial blurring, poor signal-to-noise ratio, non-ideal illumination, and the spectral response of the sensor. Following the inference of the deep network acting on the input smartphone microscope image, several spatial details were restored as illustrated Figs. 3(d) and 3(g). In addition, the deep network corrected the severe colour distortion of the smartphone image, restoring the original colours of the dyes that were used to stain the lung tissue sample, which is highly important for telepathology and related applications. Another advantage of applying the deep network is the fact that it performs denoising of the smartphone microscope images, while retaining the fidelity of the fine-resolution features, as demonstrated in Fig. 3(i). These results were also quantitatively evaluated by using the structural similarity (SSIM) index[25] calculated against the gold standard images, revealing the improvement of the neural network output images as shown in Table 2. Furthermore, as detailed in Table 3, we have applied the CIE-94 colour distance developed by the Commission internationale de l'éclairage (CIE)[26,27] as another metric to quantify the reconstruction quality of the deep network, with respect to the gold standard benchtop microscope images of the same samples. Overall, the deep network has significantly improved the average CIE-94 colour distance of the mobile microscope images by a factor of 4-11 fold, where the improvement was sample dependent as shown in Table 3. This colour improvement is especially significant for pathology field, where different dyes are used to stain various tissue structures, containing critical information for expert diagnosticians.

Using the same Masson's-trichrome-stained lung tissue data, we also evaluated the ability of the same neural network to enhance smartphone microscope images that were further degraded by applying lossy compression to them. One important advantage of applying lossy (e.g., JPEG) compression to smartphone microscope images is that compression makes them ideal for storage and transmission/sharing via the bandwidth restrictions of resource-limited environments; this also means that the deep network can perform image enhancement on demand at e.g., the office of a remote pathologist or medical expert. For the smartphone microscope images of the lung tissue, applying JPEG compression reduced an average image with a ~0.1 mm$^2$ FOV from 1.846 MB to 0.086 MB, resulting in image files that are >21 times smaller. However, lossy compression creates artefacts, such as blocking, and increases the noise and colour distortions. As demonstrated in Fig. 4, following the training of the deep network with JPEG-compressed images (see Table 1), it inferred images comparable in quality to those inferred by the deep network that was trained with lossless compression (TIFF) images. The difference was also assessed using the SSIM and the CIE-94 colour distance metrics. As summarized in Tables 2 and 3, the average SSIM was reduced by approximately 0.02, while the average CIE-94 colour distance was reduced by approximately 0.067 for the aberration corrected images, which form a negligible compromise when scenarios with strict transmission bandwidth and storage limits are considered.

Next, we applied our deep network approach to images of Pap smear samples acquired with our mobile-phone microscope (see Table 1 for implementation details). A Pap smear test is an efficient means of cervical cancer screening, and the sample slide preparation, including its staining, can be performed in a field setting, where a mobile microscope can be of great importance. Due to the thickness of the Pap smear cells (~10–15 μm), imaging such a sample using a high-NA objective lens with a shallow DOF often requires focusing on multiple sample planes. In our training procedure, we used images from a *single* plane that were acquired by automatic focusing of the benchtop microscope (see Methods section). As demonstrated in Fig. 5, the deep network, using the smartphone microscope input images, created enhanced, colour-corrected, denoised images with an extended DOF, compared to the images that were captured using the higher-NA objective lens of the benchtop microscope, also emphasized by the yellow arrows in Fig. 5.

Finally, we tested the ability of our method to increase the visibility and detection of *Trypanosoma (T.) brucei brucei* in infected blood smear samples. *T. brucei brucei* is a model organism of human African trypanosomiasis[28], which is found in 37 African countries[29] and whose corresponding parasites (i.e., *T. b. rhodesiense* and *T. b. gambiense*) put millions of people at risk of life-threatening infections[29]. The conventional method for diagnosis of acute infection caused by these parasites is imaging of thick or thin blood smear slides prepared with Giemsa stain[30] labelling the trypanosomes. The deep network inference results for a thin blood smear sample (prepared using an infected mouse blood) that is imaged by a smartphone microscope are shown in Fig. 6. The output of the deep network (Figs. 6(b) and 6(e)) demonstrates clear improvements of the smartphone microscope image, thereby revealing the sub-cellular structure and the stained nuclei of the parasites, also emphasized by the yellow arrows in Figs. 6(d,e,f). The image enhancement enabled by the deep network promotes the detectability of the trypanosomes in a blood smear sample, as shown in the comparison to the corresponding image obtained by a benchtop microscope.

**Discussion**

While our deep networks were trained with sample-specific datasets in this study, it is possible to train a universal network, at the expense of increasing the complexity of the deep network (for example, increasing the number of channels), which will accordingly increase the inference time and memory resources used[22]. This, however, is not expected to create a bottleneck since image upsampling occurs only in the last two layers in our deep network architecture. Stated differently, the upsampling process is optimized through supervised learning in this approach. Quite importantly, this design choice enables the network operations to be performed in the low-resolution image space, which reduces the time and memory requirements compared with those designs in which interpolated images are used as inputs (to match the size of the outputs)[31]. This design significantly decreases both the training and testing times and relaxes the computational resource requirements, which is important for implementation in resource-limited settings and could pave the way for future implementations running on smartphones.

In this work, the smartphone microscope images were captured using the automatic image-capture settings of the phone, which inevitably led the colour response of the sensor to be non-uniform among the acquired images. Training the deep network with such a diverse set of images creates a more robust network that will not over-fit when specific kinds of illumination and colour responses are present. In other words, the networks that we trained produced generalized, colour-corrected responses, regardless of the specific colour response acquired by using the automatic settings of the smartphone and the state of the battery-powered illumination component of the mobile microscope. This property should be very useful in actual field settings, as it will make the imaging process more user-friendly and mitigate illumination and image acquisition related variations that could become prominent when reduced energy is stored in the batteries of the illumination module. Furthermore, in recent years, the vast use of digital pathology has highlighted the differences of whole slide pathology images obtained at different laboratories due to the variability in sample preparation, staining procedures, and microscopic image scanning[32]. These variances in colour accuracy, resolution, contrast, and dynamic range of the acquired images affect the "fitness for purpose" for diagnostic use, by human observers or automated image analysis algorithms[32]. These issues have created an urgent need for optical image standardization, to better take into account such variations in different stages of the sample preparation, staining as well as imaging[32]. We believe that the presented deep learning based approach, with further training, can also be used as part of such an image standardization protocol, by transforming different microscopic images to have similar statistical properties even though they are generated at different laboratories with varying imaging platforms and staining procedures. This would help standardize the images obtained by various cost-effective and mobile microscopes, further enhance their spread and use in biomedical and clinical applications, and reduce diagnostic discrepancies that might result due to above discussed variations in the raw acquired images.

Although smartphone microscopes possess certain advantages, such as integration with off-the-shelf consumer products benefiting from economies of scale, portability, and inherent data communication, a plethora of other devices and platforms (e.g., Raspberry Pi) with different capabilities can be employed as cost-effective microscopes and benefit from the presented deep learning based approach. For example, by using a compact benchtop microscope composed of cost-effective objective lenses and illumination sources, some of the mechanical (e.g., related to object holder and its alignment) and illumination instabilities should produce less degradation in image quality than that resulting from using a smartphone-based mobile microscope. Such an imaging apparatus with its better repeatability in imaging samples will facilitate the use of the pyramid elastic registration as part of the image enhancement workflow, since the image distortions will be more stationary and less affected by mechanical and illumination instabilities resulting from, e.g., user variability and the status of the battery. For that, we could use the average block-shift correction maps calculated between the high-end and cost-effective microscope images; for example, see the mean shift map calculated for the FOV of the lung tissue sample (Supplementary Fig. 3).

To conclude, we demonstrated the significant enhancement of low-resolution, noisy, distorted images of various specimens acquired by a cost-effective, smartphone-based microscope by using a deep learning approach. This enhancement was achieved by training a deep convolutional neural network using the smartphone microscope images and corresponding benchtop microscope images of various specimens, used as gold standard. The results, which were obtained using a single feed-forward algorithm, exhibited important advantages such as the enhancement and restoration of fine spatial features, correction for the colour aberrations, and removal of noise artefacts and warping, introduced by the mobile phone microscope optical hardware/components. For samples that naturally include height/depth variations, such as Pap smear samples, we also observed the advantage of DOF extension with respect to the images of a benchtop microscope with a higher NA. These results demonstrate the potential of using smartphone-based microscopes along with deep learning to obtain high-quality images for telepathology applications, relaxing the need for bulky and expensive microscopy equipment in resource-limited settings. Finally, this presented approach might also provide the basis for a much-needed framework for standardization of optical images for clinical and biomedical applications.

**Methods**

**Design of the smartphone-based microscope**

We used a Nokia Lumia 1020 in the design of our smartphone-based transmission microscope. The regular camera application of the smartphone facilitates the capture of images in raw format (i.e., DNG) as well as JPG images using the rear camera of the smartphone, which has 41 megapixels. The same application also provides adjustable parameters such as the sensor's sensitivity (International Organization for Standardization, ISO) and exposure time. While capturing images, we set the ISO to 100, exposure time and focus to auto, and white balance to cloud modes, respectively.

Autodesk Inventor was used to design the 3D layout of the optomechanical attachment unit that transforms the smartphone into a field-portable and cost-effective microscope. It includes an *xyz* stage that facilitates lateral scanning and axial focusing. The optomechanical parts of the unit were printed using a 3D printer (Stratasys, Dimension Elite) and acrylonitrile butadiene styrene (ABS).

To provide bright-field illumination, a 12 RGB LED ring structure (NeoPixel Ring) with integrated drivers (product no. 1643) and its microcontroller (product no. 1501) were purchased from Adafruit (New York City, NY, USA). The LEDs in the ring were programmed using Arduino to provide white light to illuminate the samples. The LEDs were powered using a rechargeable battery (product no. B00EVVDZYM, Amazon, Seattle, WA, USA). The illumination unit illuminated each sample from the back side through a polymer diffuser (Zenith Polymer® diffuser, 50% transmission, 100 μm thickness, product no. SG 3201, American Optic Supply, Golden, CO, USA), as detailed in Supplementary Fig. 1. An external lens with a focal length of 2.6 mm, provided a magnification of ~2.77, a FOV of ~1 mm$^2$, and a half-pitch lateral resolution of ~0.87 μm, as demonstrated in Supplementary Fig. 1. We used the *xy* stage on the sample tray to move each sample slide for lateral scanning and the *z* stage to adjust the depth of focus of the image.

**Benchtop microscope imaging**

Image data acquisition was performed using an Olympus IX83 microscope equipped with a motorized stage. The images were acquired using a set of Super Apochromat objectives, (Olympus UPLSAPO 20X/0.75NA, WD0.65). The colour images were obtained using a Qimaging Retiga 4000R camera with a pixel size of 7.4 μm. The microscope was controlled by MetaMorph® microscope automation software (Molecular Devices, LLC), which includes automatic slide scanning with autofocusing. The samples were illuminated using a 0.55NA condenser (Olympus IX2-LWUCD).

**Sample preparation**

All of the human samples were obtained after de-identification of the patients and related information and were prepared from existing specimens. Therefore, this work did not interfere with the standard care practices or sample collection procedures.

*Lung tissue*: De-identified formalin-fixed paraffin-embedded Masson's-trichrome-stained lung tissue sections from two patients were obtained from the Translational Pathology Core Laboratory at UCLA. The samples were stained at the Histology Lab at UCLA.

*Pap smear:* A de-identified Pap smear slide was provided by UCLA Department of Pathology.

*Trypanosome infected mouse blood smear*: To stain the blood smear sample, we used Fisher HealthCare™ PROTOCOL™ Hema 3™ Solutions (product no. 22-122911) purchased from Fisher Scientific (Hampton, NH, USA). Mouse blood infected with *T. brucei brucei* at a parasitemia level of $3.3 \times 10^7$ trypanosomes/mL was provided by the laboratory of Prof. Kent Hill at the UCLA Microbiology, Immunology, and Molecular Genetics Department. To make the blood smear sample, 4 μL of trypanosome-infected mouse blood was put on a clean glass microscope slide. The drop of blood was spread using another clean microscope slide at 45° angle. After the thin film of blood had completely dried, it was fixed using absolute methanol and allowed to dry completely. We dipped the slide five times for 1 s each in Hema 3 Solution I and allowed the excess solution to drain. The last step was repeated using Hema 3 Solution II. The slide was then rinsed and allowed to dry for imaging.

**Data preprocessing**

To ensure that the deep network learns to enhance smartphone microscope images, it is important to preprocess the training image data so that the smartphone and benchtop microscope images will match. The deep network learns how to enhance the images by following an accurate smartphone and benchtop microscope FOV matching process, which in our designed network is based on a series of spatial operators (convolution kernels). Providing the deep network with accurately registered training image data enables the network to focus the learning process on correcting for repeated patterns of distortions between the images (input vs. gold standard), making the network more compact and resilient overall and requiring less data and time for training and data inference.

This image registration task is divided into two parts. The first part matches the FOV of an image acquired using the smartphone microscope with that of an image captured using the benchtop microscope. This FOV matching procedure can be described as follows: (i) Each cell phone image is converted from DNG format into TIFF (or JPEG) format with the central 0.685 $mm^2$ FOV being cropped into four parts, each with 1024×1024 pixels. (ii) Large-FOV, high-resolution benchtop microscope images (~25K×25K pixels) are formed by stitching 2048×2048 pixel benchtop microscope images. (iii) These large-FOV images and the smartphone image are used as inputs for scale-invariant feature transform (SIFT)[33] and random sample consensus (RANSAC) algorithms. First, both colour images are converted into grey-scale images. Then, the SIFT frames (*F*) and SIFT descriptors (*D*) of the two images are computed. *F* is a feature frame and contains the fractional centre of the frame, scale, and orientation. *D* is the descriptor of the corresponding frame in *F*. The two sets of SIFT descriptors are then matched to determine the index of the best match. (iv) A homography matrix, computed using RANSAC, is used to project the low-resolution smartphone image to match the FOV of the high-resolution benchtop microscope image, used as gold standard.

Following this FOV matching procedure, the smartphone and benchtop microscope images are globally matched. However, they are not accurately registered, mainly due to distortions caused by the imperfections of the optical components used in the smartphone microscope design and inaccuracies originating during the mechanical scanning of the sample slide using the *xyz* translation stage. This second part of the registration process locally corrects for all these distortions between the input and gold standard images by applying a pyramid elastic registration algorithm, which is depicted in Supplementary Fig. 2. During each iteration of this algorithm, both the smartphone and corresponding benchtop microscope images are divided into $N \times N$ blocks, where typically $N = 5$. A block-wise cross-correlation is calculated using the corresponding blocks from the two images. The peak location inside each block represents the shift of its centre. The peak value, i.e., the Pearson correlation coefficient[34], represents the similarity of the two blocks. A cross-correlation map (CCM) and an $N \times N$ similarity map are extracted by locating the peak locations and fitting their values. An $m \times n$ translation map is then generated based on the weighted average of the CCM at each pixel. This translation map defines a linear transform from the distorted image to the target enhanced image. This translation operation, although it corrects distortions to a certain degree, is synthesized from the block-averaged CCM and therefore should be refined with smaller-block-size CCMs. In the next iteration, $N$ is increased from 5 to 7, and the block size is reduced. This iterative procedure is repeated until the minimum block size is reached, which we empirically set to be $m \times n = 50 \times 50$ pixels. The elastic registration in each loop followed the open-source NanoJ plugin in ImageJ[35,36].

Following the FOV matching and registration steps discussed above, the last step is to upsample the target image in a way that will enable the network to learn the statistical transformation from the low-resolution smartphone images into high-resolution, benchtop-microscope equivalent images. When the benchtop microscope was used to create gold standard images used for training, each sample was illuminated using a 0.55NA condenser, which creates a theoretical resolution limit of approximately 0.4 µm using a 0.75 NA objective lens (20X). However, the lateral resolution is constrained by the effective pixel size at the CCD, which is 7.4 µm; therefore the practical half-pitch resolution of the benchtop microscope using a 20X objective lens is: 7.4 µm/20 = 0.37 µm, corresponding to a period of 0.74 µm. On the other hand, the smartphone microscope is based on a CMOS imager and has a half-pitch resolution of 0.87 µm, corresponding to a resolvable period of 1.74 µm (Supplementary Fig. 1). Thus, the desired upsampling ratio between the smartphone and benchtop microscope images is given by 0.87/0.37 = 2.35. Therefore, we trained the deep network to upsample by a ratio of 2.5, and by applying the upsampling only at the final convolutional layers, we enabled the network structure to remain compact, making it easier to train and infer [21,31].

**Deep neural network architecture and implementation**

The deep neural network architecture follows our previous design[21]. The network receives three input feature maps (RGB channels); then, following the first convolutional layer, the number of feature maps is expanded to 32. Formally, the convolution operator of the $i$-th convolutional layer for $x,y$-th pixel in the $j$-th feature map is given by:

$$g_{i,j}^{x,y} = \sum_{r}\sum_{u=0}^{U-1}\sum_{v=0}^{V-1} w_{i,j,r}^{u,v} g_{i-1,r}^{x+u,y+v} + b_{i,j}, \quad (1)$$

where $g$ defines the feature maps (input and output), $b_{i,j}$ is a learned bias term, $r$ is the index of the feature maps in the convolutional layer, and $w_{i,j,r}^{u,v}$ is the learned convolution kernel value at its $u,v$-th entry. The size of the convolutional kernel is $U \times V$, which we set to be 3×3 throughout the network. Following the initial expansion of the number of feature maps from 3 to 32, the network consists of five residual blocks, which contribute to the improved training and convergence speed of our deep networks[13]. The residual blocks implement the following structure:

$$X_{k+1} = X_k + \text{ReLU}(\text{Conv}_{k\_2}(\text{ReLU}(\text{Conv}_{k\_1}(X_k)))), \quad (2)$$

where Conv(.) is the operator of each convolutional layer (equation (1)), and the non-linear activation function that we applied throughout the deep network is ReLU, defined as $\text{ReLu}(x) = \max(0, x)$. The number of feature maps for the $k$-th residual block is given by[37]

$$A_k = A_{k-1} + \text{floor}((\alpha \times k) / K + 0.5), \quad (3)$$

where $K = 5$ is the total number of residual blocks, $k = [1:5]$, $\alpha = 10$, and $A_0 = 32$. By gradually increasing the number of feature maps throughout the deep network (instead of having a constant large number of feature maps), we keep the network more compact and less demanding on computational resources (for both training and inference). However, increasing the number of channels through residual connections creates a dimensional mismatch between the features represented by $X_k$ and $X_{k+1}$ in equation (2). To avoid this issue, we augmented $X_k$ with zero-valued feature maps, to match the total number of feature maps in $X_{k+1}$. Following the output of the fifth residual block, another convolutional layer increases the number of feature maps from 62 to 75. The following two layers transform these 75 feature maps, each with $S \times T$ pixels, into three output channels, each with $(S \times L) \times (T \times L)$ pixels, which correspond to the RGB channels of the target image. In our case, we set $L$ to 2.5 (as detailed in the Data Preprocessing section, related to upsampling). To summarize, the number of feature maps in the convolutional layers in our deep network follows the sequence of: 3 → 32 → 32 → 34 → 34 → 38 → 38 → 44 → 44 → 52 → 52 → 62 → 62 → 75 → 3 → 3. If the number of pixels in the input is odd, the size of the output is given by $3 \times \lceil (S \times L) \rceil \times \lceil (T \times L) \rceil$. Performing upsampling only at the final layers further reduces the computational complexity, increases the training and inference speed, and enables the deep network to learn an optimal upsampling operator.

The network was trained to optimize the loss function $\ell$ based on the current network output $Y^{\Theta} = \Phi(X_{input}; \Theta)$ and the target (benchtop microscope) image $Y^{Label}$:

$$\ell(\Theta) = \frac{1}{3 \times S \times T \times L^2} \left[ \sum_{c=1}^{3} \sum_{s=1}^{S \times L} \sum_{t=1}^{T \times L} \left\| Y^{\Theta}_{c,s,t} - Y^{Label}_{c,s,t} \right\|_2^2 + \lambda \sum_{c=1}^{3} \sum_{s=1}^{S \times L} \sum_{t=1}^{T \times L} \left| \nabla Y^{\Theta} \right|^2_{c,s,t} \right], \quad (4)$$

where $X_{input}$ is the network input (smartphone microscope raw image), with the deep network operator denoted as $\Phi$ and the trainable network parameter space as $\Theta$. The indices $c$, $s$, and $t$ denote the $s,t$-th pixel of the $c$-th colour channel. The loss function (equation (4)) balances the mean-squared error and image sharpness with a regularization parameter $\lambda$, which was set to be 0.001. The sharpness term, $\left|\nabla Y^{\Theta}\right|^2_{c,s,t}$ is defined as[38] $\left|\nabla Y^{\Theta}\right|^2 = \left(h * Y^{\Theta}\right)^2 + \left(h^T * Y^{\Theta}\right)^2$, where

$$h = \begin{bmatrix} -1 & 0 & 1 \\ -2 & 0 & 2 \\ -1 & 0 & 1 \end{bmatrix}, \quad (5)$$

and $(.)^T$ is the matrix transpose operator.

The calculated loss function is then back-propagated to update the network parameters ($\Theta$), by applying the adaptive moment estimation optimizer[39] with a constant learning rate of $2\times10^{-4}$. During the training stage, we chose to train the network with a mini-batch of 32 patches (see Table 1). The convolution kernels were initialized by using a truncated normal distribution with a standard deviation of 0.05 and a mean of $0^{13}$. All the network biases were initialized as 0.

**Colour distance calculations**

The average and the standard deviation of the CIE-94 were calculated between the 2.5× bicubic upsampled smartphone microscope raw *input* images and the benchtop microscope images (used as gold standard), as well as between the deep network *output* images and the corresponding benchtop microscope images, on a pixel-by-pixel basis and was averaged across the images of different samples (see Table 3). As reported in Table 3, we have also performed the CIE-94 colour difference calculations[26] on warp-corrected (using the pyramid elastic registration algorithm) and 2.5× bicubic upsampled smartphone microscope images as well as on their corresponding network output images, all calculated with respect to the same gold standard benchtop microscope images.

**Acknowledgements**

The authors acknowledge Dr. Kent Hill and Dr. Michelle Shimogawa of UCLA for providing *T.brucei brucei* infected mouse blood samples.

**Data and materials availability**: All the data and methods needed to evaluate the conclusions in this work are present in the main text and the Supplementary Information.

**Competing financial interests:** The authors declare no competing financial interests.

**Funding**: The Ozcan Research Group at UCLA acknowledges the support of NSF Engineering Research Center (ERC, PATHS-UP), the Army Research Office (ARO; W911NF-13-1-0419 and W911NF-13-1-0197), the ARO Life Sciences Division, the National Science Foundation (NSF) CBET Division Biophotonics Program, the NSF Emerging Frontiers in Research and Innovation (EFRI) Award, the NSF EAGER Award, NSF INSPIRE Award, NSF Partnerships for Innovation: Building Innovation Capacity (PFI:BIC) Program, Office of Naval Research (ONR), the National Institutes of Health (NIH), the Howard Hughes Medical Institute (HHMI), Vodafone Americas Foundation, the Mary Kay Foundation, Steven & Alexandra Cohen Foundation, and KAUST. This work is based upon research performed in a laboratory renovated by the National Science Foundation under Grant No. 0963183, which is an award funded under the American Recovery and Reinvestment Act of 2009 (ARRA). Yair Rivenson is partially supported by the European Union's Horizon 2020 research and innovation programme under the Marie Skłodowska-Curie grant agreement No H2020-MSCA-IF-2014-659595 (MCMQCT).

**Author contributions**: A.O. and Y. R. conceived the research, H.C.K. and D.T. have designed the add-on smartphone microscope unit, H.C.K., H.W., Y. R. and Z.G. conducted the experiments, Y.R., H.W., Z.W., Z. R., H. G., Y. Z., and K. L. processed the data. Y.R., H.C.K, H.W., and A.O. prepared the manuscript and all the other authors contributed to the manuscript. A.O. supervised the research.

**Figures**

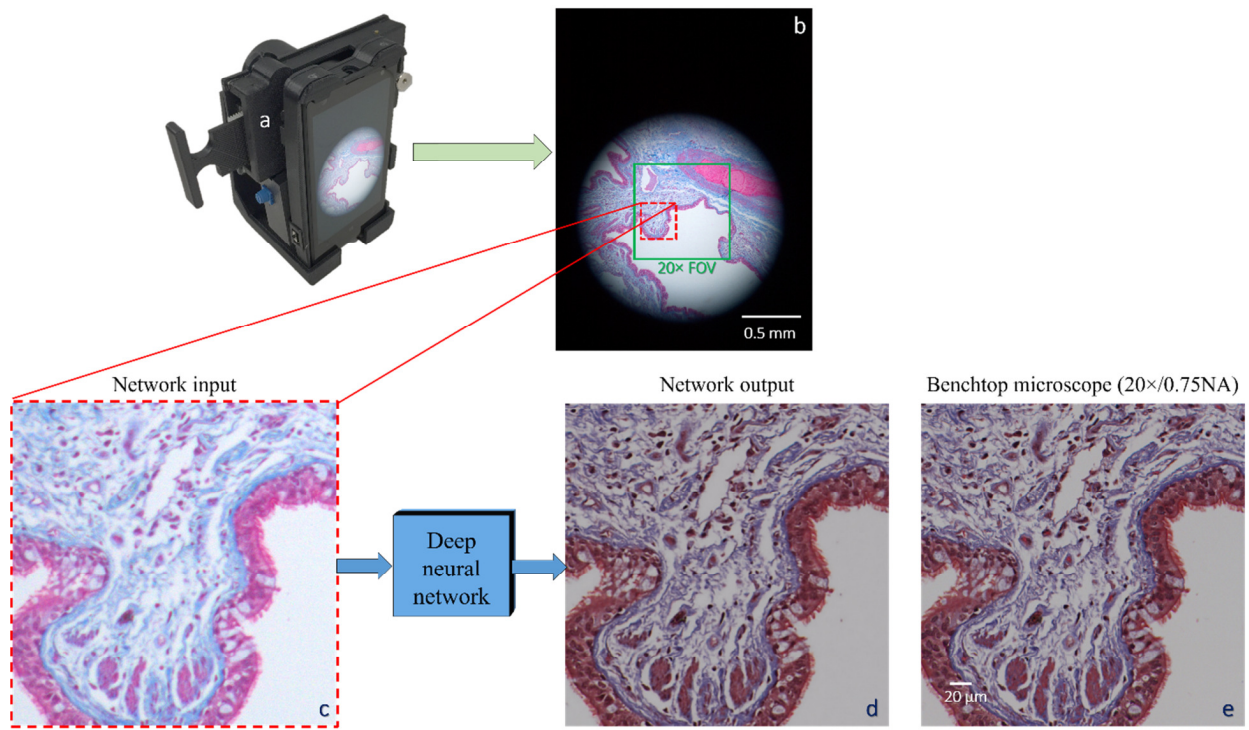

**Fig. 1**. (a, b) Masson's-trichrome-stained lung tissue sample image acquisition using a cost-effective smartphone microscope device. (c) Input region of interest (ROI), for which the deep network blindly yields (d) an improved output image, which resembles (e) an image obtained using a high-end benchtop microscope, equipped with a 20×/0.75NA objective lens and a 0.55NA condenser.

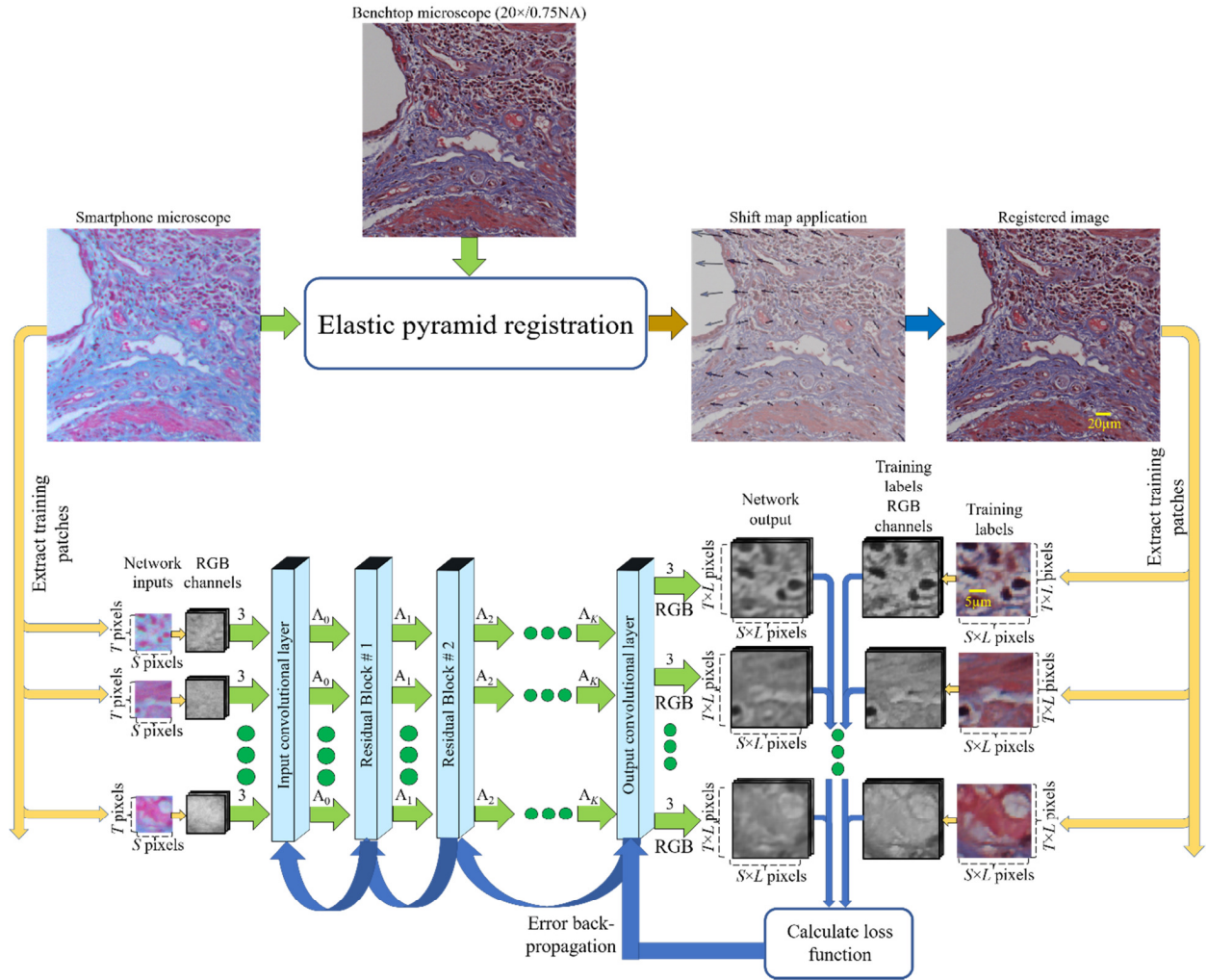

**Fig. 2**. Training phase of the deep neural network.

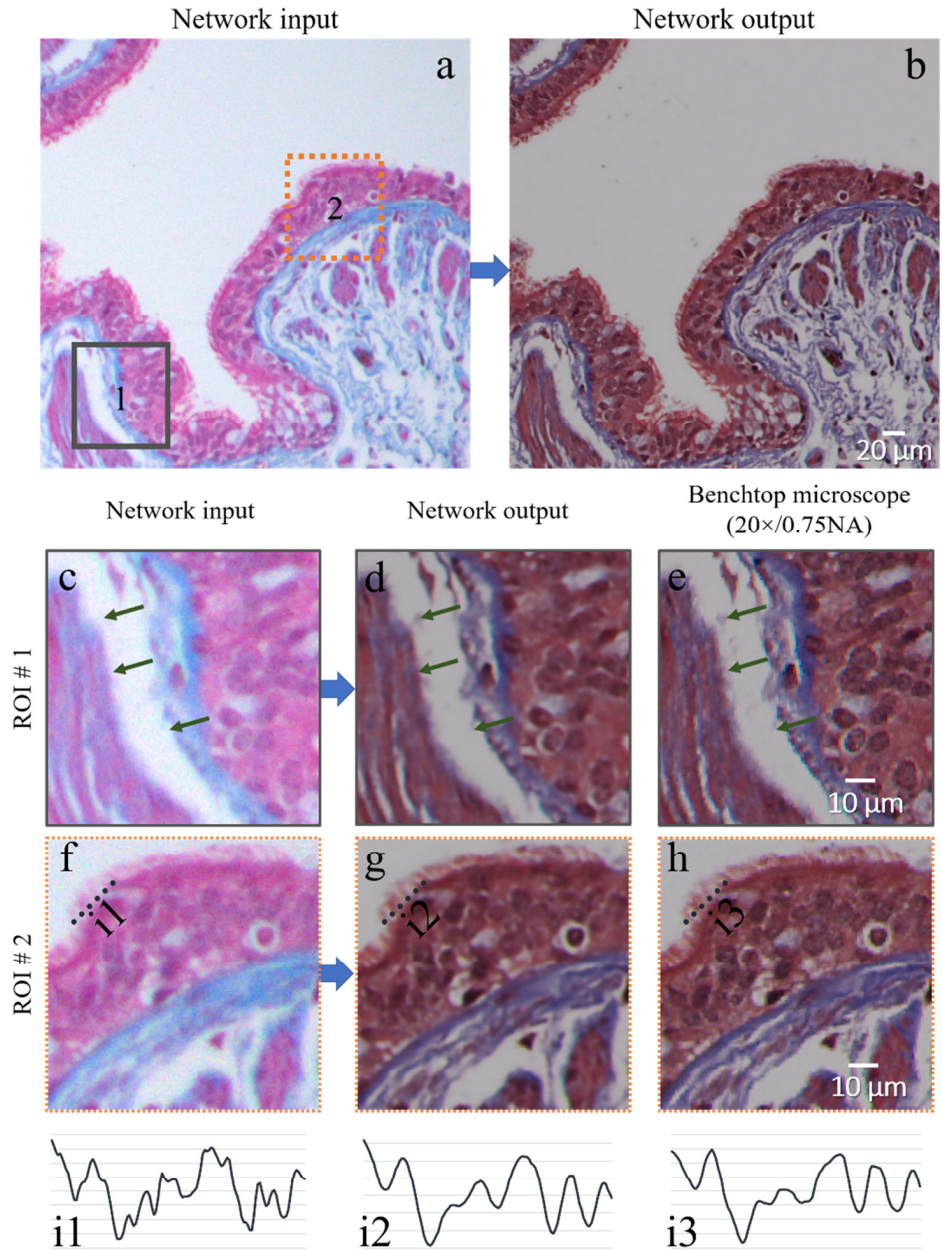

**Fig. 3**. Deep neural network output for a Masson's-trichrome-stained lung tissue section. (a) Smartphone microscope image, and (b) its corresponding deep network output. Zoomed-in versions of the ROIs shown in (c, f) the smartphone input image and (d, g) the neural network output image. (e, h) Images of the same ROIs acquired using a 20×/0.75NA objective lens (with a 0.55NA condenser). The green arrows in (c, d, e) point to some examples of the fine structural details that were recovered using the deep network. Several other examples can be found in (d, g) compared to (c, f), which altogether highlight the significant improvements in the deep network output images, revealing the fine spatial and spectral details of the sample. (i) Cross-section line profiles from (f, g, h) demonstrating the noise removal performed by the deep network, while retaining the high-resolution spatial features.

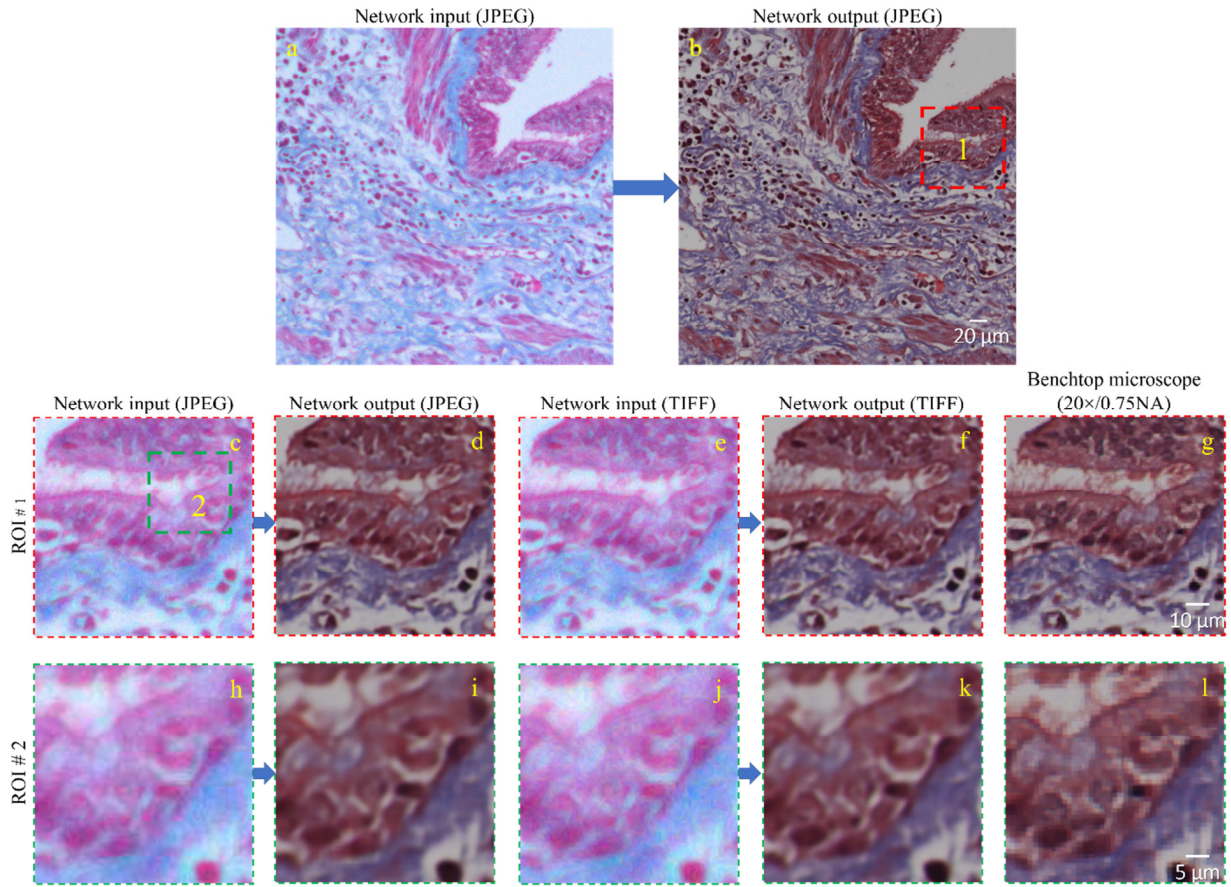

**Fig. 4**. Comparison of the deep network inference performance when trained with lossy compression (JPEG) and lossless compression (TIFF). (a) JPEG-compressed image, and (b) its corresponding deep network output. Zoomed-in versions of (c–g) ROI #1 and (h–l) ROI #2.

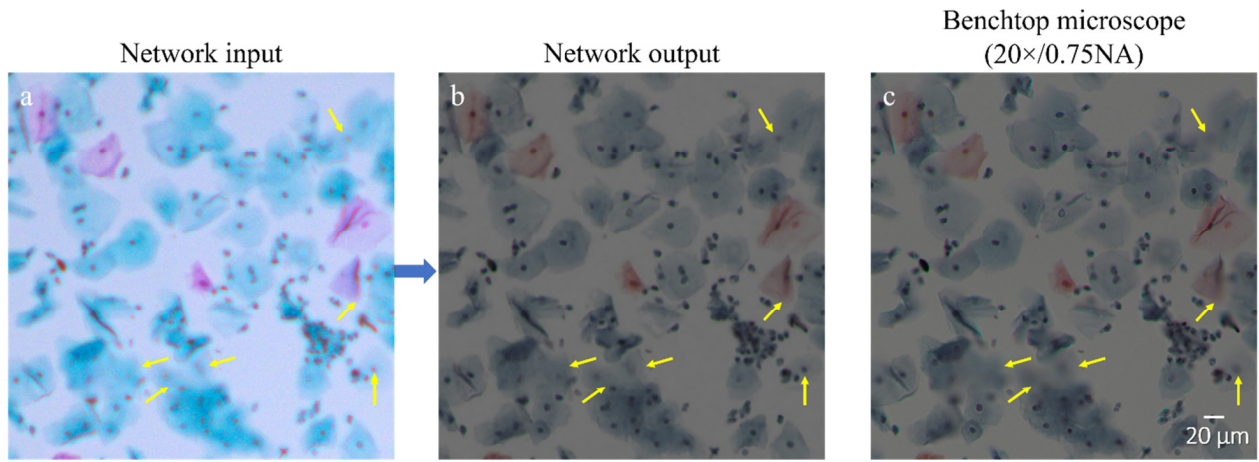

**Fig. 5**. Deep neural network output image corresponding to a stained Pap smear sample. (a) Smartphone microscope image, (b) its corresponding deep network output, and (c) a 20×/0.75NA benchtop microscope image. The yellow arrows help us reveal the extended DOF of the imaging results obtained by the deep network.

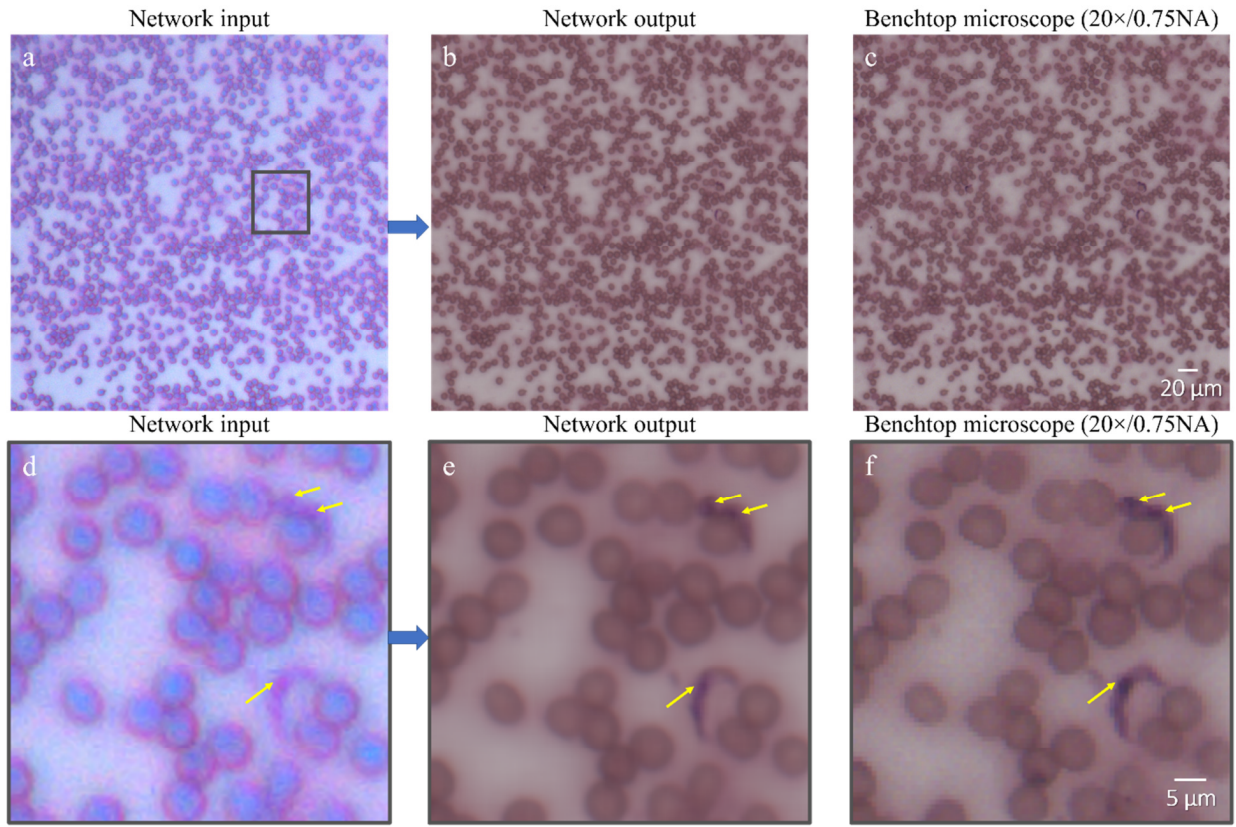

**Fig. 6**. Deep neural network output image for trypanosome-infected mouse blood smear sample. (a) Smartphone microscope image, (b) its corresponding deep network output, and (c) a 20×/0.75NA benchtop microscope image. (d) Zoomed-in version of the ROI of the smartphone input image, (e) corresponding network output, and (f) 20×/0.75NA benchtop microscope image, revealing the image enhancement achieved by the deep neural network. The yellow arrows point to features (i.e., stained trypanosome nuclei) that were accurately coloured and resolved by the deep network, despite having very low visibility in the raw smartphone image.

**Tables**

|  | Number of input–output patches (number of pixels in each mobile-phone microscope image) | Validation set (number of pixels in each mobile-phone microscope image) | Number of epochs till convergence | Training time |
|---|---|---|---|---|
| Masson's-trichrome-stained lung tissue | 129,472 patches (60×60 pixels) | 95 images (800×800 pixels) | 134 | 36 h, 40 min |
| H&E-stained Pap smear | 222,008 patches (60×60 pixels) | 63 images (1024×1024 pixels) | 190 | 20 h, 24 min |
| Blood smear | 43,528 patches (60×60 pixels) | 17 images (1024×1024 pixels) | 324 | 8 h, 42 min |

**Table 1.** Deep neural network training details for different samples.

|  | Test set | Bicubic upsampling SSIM | Deep neural network SSIM |
|---|---|---|---|
| Masson's-trichrome-stained lung tissue (TIFF input) | 90 images (800×800 pixels) | 0.4956 | 0.7020 |
| Masson's-trichrome-stained lung tissue (JPEG input) | 90 images (800×800 pixels) | 0.5420 | 0.6830 |
| H&E-stained Pap smear | 64 images (1024×1024 pixels) | 0.4601 | 0.7775 |
| Blood smear | 15 images (1024×1024 pixels) | 0.6605 | 0.8525 |

**Table 2.** Average SSIM for the different pathology samples, comparing bicubic ×2.5 upsampling of the smartphone microscope images and the deep neural network output images.

|  | (A) Raw smartphone microscope images | | (B) Warp-corrected smartphone microscope images | | (C) Deep network output images of (A) | | (D) Deep network output images of (B) | |
| --- | --- | --- | --- | --- | --- | --- | --- | --- |
|  | Average | Std | Average | Std | Average | Std | Average | Std |
| Masson's-trichrome-stained lung tissue (TIFF) | 15.976 | 1.709 | 16.077 | 1.683 | 4.369 | 0.917 | 3.814 | 0.797 |
| Masson's-trichrome-stained lung tissue (JPEG) | 15.914 | 1.722 | 15.063 | 1.820 | 4.372 | 0.847 | 3.747 | 0.908 |
| H&E-stained Pap smear (TIFF) | 26.230 | 0.766 | 23.725 | 0.969 | 2.127 | 0.267 | 2.092 | 0.317 |
| Blood smear (TIFF) | 17.310 | 0.726 | 16.884 | 0.664 | 2.143 | 0.452 | 1.431 | 0.099 |

**Table 3.** Average and standard deviation (Std) of the CIE-94 colour distances compared to the gold standard benchtop microscope images for the different pathology samples.